# Detecting Opinions in Tweets

Abdelmalek Amine[1*], Reda Mohamed Hamou[1], Michel Simonet[2]

[1]GeCoDe Laboratory, Department of Computer Science, Tahar Moulay University, Saida, Algeria

[2]AGIM (AGe, Imagerie, Modélisation), Faculté de Médecine de Grenoble, Grenoble, France

**ABSTRACT**

Given the incessant growth of documents describing the opinions of different people circulating on the web, including Web 2.0 has made it possible to give an opinion on any product in the net. In this paper, we examine the various opinions expressed in the tweets and classify them (positive, negative or neutral) by using the emoticons for the Bayesian method and adjectives and adverbs for the Turney's method.

**Keywords:** Opinions detection, Tweets, Supervised approaches, Semantic orientation, Classification

## INTRODUCTION

Nowadays, we live in a world where information is available in large quantities and are of very different qualities. Internet has continued to evolve with new content. For example, companies store increasing quantity of data. Email is an extremely popular means of communication. Documents that were once printed as manuscripts are now available in digital format. However, all these complex information would be meaningless if we cannot easily access them. So, we need tools to search, sort, store, update and analyse these available data. It is therefore necessary to provide systems to access the desired information as quickly as possible, thus reducing human involvement.

Thousands of documents are easily available through the networks and the data-processing supports, of which more than 90% are textual documents; therefore, it becomes difficult to access these information without the support of specific tools. The text mining, which is a specialisation of data mining, answers these problems. The study of text mining is based particularly on very close links between research in textual linguistics and on adequate formalisations for a data-processing realisation. The text mining is known for its power to extract key concepts from textual information sources.

With the appearance of Web 2.0, more and more textual documents containing information that express opinions or feelings are available (comments in the social networks, comments for the evaluation of products by customers, forums, newsgroup and blogs).

The automatic extraction of opinions (the opinion mining), which is a specialisation of text mining, is a field of research in full rise. It becomes essential for several fields of applications.

Opinion mining indicates the methods of identification of the opinions and argumentations within a set of texts and, among these known applications, we note in particular the text clustering.

Our study is focused on detection of opinions in the tweets (textual information available on Twitter) by using some methods of classification of texts.

Several methods of classification were developed, each having its advantages and disadvantages. The first goal of our research was to examine the various opinions in the tweets, i.e. to classify them (positive, negative or neutral) by using the emoticons for the Bayesian method and the adjectives and the adverbs for the Turney's method.

**STATE-OF-THE-ART**

The analysis of text in terms of study of the feelings, opinions or points-of-view is not recent [2], [15]. However, the field of the excavation of opinions and the analysis of the feelings took an important position since 2000 with the arrival of the community web and the multiplication of the forums on the net. Since then, opinion mining has become a major stake for every company wishing better comprehension of the likes and dislikes of its customers, relying on the opinions of those who wish to compare the products before buying them. For example, Morinaga *et al*. [11] explain how they check the reputation of the target products by analysing criticisms of the customers. First of all they access the web pages discussing about the concerned product and extract the sentences that express the opinion. They then classify the sentences according to whether they express a negative opinion or a positive opinion and deduce the popularity of the product. In this context, Turney[13] has classified comments into two categories, namely, recommended and not recommended.

Wilson *et al*. [16] append this classification according to popularity, the force of the expressed opinion. A lot of studies were carried out on the subject and three main categories of methods were described, namely, the approaches based on the linguistics, the approaches based on machine learning and the hybrid approaches.

CREATION OF THE LEXICONS OF OPINION

This method requires the construction of one or several lexicons of opinion. To build such lexicons, three methods can be applied as given below:

*Manual Method*

It consists of filling the lexicon with words of opinions without any assistance from any particular tool. The selection of the words carrying opinion and the choice of their polarity are thus done only by human expertise. We can suppose that a considerable part of subjectivity may affect the process and involve errors in classification. Whatever the method of construction of the lexicons, a first stage of manual classification is necessary.

*Method based on the Corpora*

It completes the lexicon of opinions and consists of using the coordinating conjunctions present between a word already classified and an unclassified word[4], [5], [8]. For example, if a conjunction AND separates a positive classified word in the lexicon of opinions and an unclassified word, then the unclassified word will be regarded as being positive. On the contrary, if the conjunction BUT separates a positive classified word and a unclassified word, then the unclassified word will be regarded as being negative. The conjunctions used are as follows: AND, OR, BUT, EITHER-OR and NEITHER-NOR.

*Method based on the Dictionaries*

It consists of using existing dictionaries of synonyms and antonyms, such as WordNet[10], in order to determine the semantic orientation (SO) of new words. For example, Hu and Liu[7] use this dictionary in order to predict the SO of the adjectives. In WordNet, the words are organised like trees.

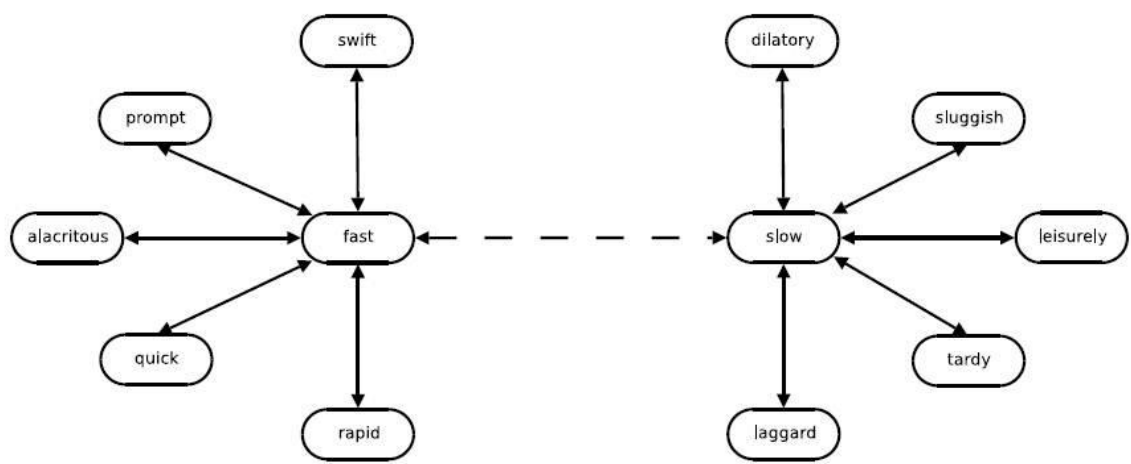

**Figure 1**: Example of tree of synonyms and antonyms in WordNet (full arrow = synonymous, hatched arrow = antonyms)

APPROACHES BASED ON MACHINE LEARNING

In these types of approaches, the words are usually regarded as equivalent variables. Thus, the semantic aspect is not considered. The methods mostly used for classification of opinion are those of supervised classification. These types of methods consist of building a model of classification using samples. Samples are data whose class we already know. In this case, we refer to classified or labelled data.

HYBRID APPROACHES

Several types of hybridisation **methods** are discussed in the literature. They all use elements described in the two previous approaches. Among these approaches, we can distinguish three different methods, namely:

- ✓ Linguistics with the service of machine learning;
- ✓ Machine learning in the service of linguistics;
- ✓ A fusion a posteriori of the results of the two approaches.

**PROPOSED APPROACH**

In our approach, we used two techniques, namely, supervised training represented by the probabilistic Bayesian model and SO represented by the algorithm of Turney.

SUPERVISED TRAINING (BAYESIAN ALGORITHM)

In an abstract manner, the probabilistic model for a Bayesian classifier is a conditional model. It is based on the rule of Bayes which is as follows:

$$P(A|B_1, B_2, ..., B_n) = \frac{P(B_1, ..., B_n|A) * P(A)}{P(B_1, ..., B_n)}$$

The probability of having event *A* being given $B_1$, ......., $B_n$ is given by the relationship between the probability of having the $B_1$, ......, $B_n$ events being given *A* and the probability that $B_1$,..........., $B_n$ occurred. As long as the denominator does not depend on event *A*, we can consider the probability $P(B_1... B_n)$ as being constant. The numerator can still be written in the following way:

$$P(B_1, ..., B_n|A) * P(A)$$

$$\begin{aligned}
&= P(A, B_1, ..., B_n) \\
&= P(A) * P(B_1|A) * P(B_2, ..., B_n|A, B_1) \\
&= P(A) * P(B_1|A) * P(B_2|A, B_1) \\
&\quad * P(B_3, ..., B_n|A, B_1, B_2)
\end{aligned}$$

The decomposition of $P(A, B_1... B_n)$ finishes when we traverse the whole of the $B_1 ... B_n$ classes. The 'naïve' character of this theorem is due to the fact that we suppose the independence of the various classes $B_i... B_j$ which results in:

$$P(B_i|A, B_j) = P(B_i|A)$$

This assumption also permits us to write

$$\begin{aligned}
P(A, B_1, ., B_n) &= P(A) * P(B_1|A) * .. \\
&\quad * P(B_2|A) * ... * P(B_n|A) \\
&= P(A) \prod_{i=1}^{n} P(B_i|A)
\end{aligned}$$

This theorem has many applications in the data processing field and, in particular, in speech processing, image processing, etc. We have applied this theorem for the detection of opinions in documents (tweets).

*Application to the Classification of Texts (Tweets)*

Let us illustrate the equations of the theorem seen in the preceding section with the problem of the classification of texts. Let us suppose that we have *N* categories of documents, to determine to which category $C_i$ a document *D* is associated, we calculate the membership probability of the document *D* to the category $C_i$. Based on the previously stated theorem, we can calculate this probability as follows:

$$P(C_i|D) = \frac{P(D|C_i) * P(C_i)}{P(D)}$$

In this formula, $P(C_i|D)$ represents the membership probability of the document *D* to the category $C_i$, which can also be given by evaluating the frequency of appearance of the words of the document *D* that are associated to the category $C_i$. $P(D|C_i)$ is the probability according to which, for a given category, the words of the document *D* are associated to the category $C_i$. $P(C_i)$ is the probability that associates the document *D* to the category $C_i$ independent of the contents of the document. $P(D)$ is the proper probability of the document *D*.

For really determining which category a document belongs to, it is necessary to calculate $P(C_i|D)$ for each category. Since $P(D)$ remains constant for all the categories, determination of $P(C_i|D)$ will be merely reduced to the calculation of $P(D|C_i) * P(C_i)$

BY SEMANTIC ORIENTATION (TURNEY ALGORITHM)

The algorithm takes a text written as input and produces a classification as output. The first step uses a tagger of grammatical category to identify expressions in the contributing text that contain adjectives or adverbs. The second step evaluates the SO of each extracted expression[5]. An expression has a positive SO when it has good associations (for example, 'romantic environment') and a negative SO when it has bad associations (for example, 'terrifying events')[13]. The third step assigns the given text to a class, recommended or not recommended, basing the orientation on semantic mean of the expressions extracted from the text. If the mean is positive, the prediction is that the examination recommends the article which discusses it. Otherwise, the prediction is that we do not recommend the article.

The PMI-IR algorithm (Pointwise Mutual Information - Information Retrieval) is used to assess the semantic orientation of an expression [12]. PMI-IR uses the Pointwise Mutual Information (PMI) and Information Retrieval (IR) to measure the similarity of pairs of words or expressions. The semantic orientation of a given expression is calculated by comparing its similarity to a word having a positive reference ("excellent") and subtracting the mutual information between the given expression and the word "poor".

*Algorithm Behaviour*

The first step of the algorithm extracts the expressions containing the adjectives or the adverbs. Previous works express that adjectives are good indicators of subjective evaluating sentences [6], [14].

However, although an isolated adjective can indicate subjectivity, there can be the insufficient context to determine the SO. For example, the adjective 'unforeseeable' can have a negative orientation in a car review, for example, in an expression like 'the unforeseeable direction', but it could have a positive orientation in a movie review, for example, in an expression like 'the unforeseeable plot'. Thus, the algorithm extracts two consecutive words, where a member of the pair is an adjective or an adverb and the second provides the context[13].

Initially, a tagger of grammatical category is applied to the text [1].

Two consecutive words are extracted from the text if their labels conform to any of the models given in Table 1. Labels JJ indicate adjectives, labels NN represent names, labels RB represent adverbs and labels VB represent verbs.

In the second model, two consecutive words are extracted; if the first one is an adverb and the second is an adjective, the third word (which is not extracted) cannot be a name. NNP and NNPS (of the singular and plural proper names) are avoided so that the names of the articles in the exam (the review) cannot influence the classification.

|    | First Word       | Second Word          | Third Word (Not Extracted) |
|----|------------------|----------------------|----------------------------|
| 1. | JJ               | NN or NNS            | anything                   |
| 2. | RB, RBR, or RBS  | JJ                   | not NN nor NNS             |
| 3. | JJ               | JJ                   | not NN nor NNS             |
| 4. | NN or NNS        | JJ                   | not NN nor NNS             |
| 5. | RB, RBR, or RBS  | VB, VBD, VBN, or VBG | anything                   |

**Table 1**: Models of labels to extract expressions with two words of exam (of reviews)[13].

The second step must evaluate the SO of the extracted expressions using the PMI-IR algorithm. This algorithm uses mutual information, such as measurement of the semantic force of association between the two words [3]. PMI-IR was empirically evaluated using 80 questions of test of synonym of the English foreign language test (TOEFL), obtaining a score of 74% [12]. For comparison, Latent Semantic analysis (LSA), another statistical measurement of association of words, reached a score of 64% out of same 80 questions of TOEFL [9]. Pointwise mutual informations (PMI) between two words, word1 and word2, are defined as follows [3]:

$$\text{PMI}(word_1, word_2) = \log_2 \left[ \frac{p(word_1 \,\&\, word_2)}{p(word_1)\, p(word_2)} \right] \quad (1)$$

p(*word1* and *word2*) are the probability that word1 and word2 co-occur. If the words are statistically independent, the product gives the probability that they co-occur as p(*word1*) p(*word2*). The ratio between p(*word1* and *word2*) and p(*word1*) p(*word2*) is thus a measurement of the statistical degree of dependence between the words. The log of this ratio is the quantity of information that we acquire from the presence of one of the words when we observe the other. The SO of an expression is calculated here using the equation:

$$SO(phrase) = \text{PMI}(phrase, \text{``excellent''}) - \text{PMI}(phrase, \text{``poor''}) \quad (2)$$

The reference words 'Excellent' and 'weak' were selected because, in the evaluation system of star review, it is common to define a 'weak' star and stars with 'excellent'. The SO is positive when the expression is more strongly associated to 'excellent' and negative when the expression is more strongly associated to 'weak'[13].

The PMI-IR estimates PMI by publishing questions to a search engine (from the IR in PMI-IR) and noting the success number (corresponding to documents). The following experiments use advanced AltaVista search engine, which indexes approximately 350 million web pages (only for English pages). AltaVista was chosen because it uses operator NEAR. NEAR of AltaVista allows the research of documents that contain the 10 words in different order.

$$SO(phrase) = \log_2\left[\frac{\text{hits}(phrase \text{ NEAR ``excellent''}) \text{ hits}(\text{``poor''})}{\text{hits}(phrase \text{ NEAR ``poor''}) \text{ hits}(\text{``excellent''})}\right] \quad (3)$$

To avoid division by zero, it adds 0.01 to successes (hits).

The third step must calculate the average SO of the expressions in the given text and classify the text as recommended, if the average is positive and as not recommended, if the average is negative[13].

**RESULTS AND DISCUSSIONS**

CONSTITUTION OF CORPUS

The constitution of corpus was a rather hard task, which took a lot of time. Despite the research that we have done, we could not find a corpus of tweets (all the corpora on the net are not free) and consequently we collected different tweets that come from a corpus. For this, we followed the following steps:

*Selection of Tweets*

For the collection of the tweets, we have used a java API (Application Programming Interface) called 'Twitter4J'

Twitter4J permits access to the Twitter database to recover/to post information. The API breaks up into four classes:

SEARCH: allows inquiring Twitter to recover simple data, primarily tweets.

REST: advanced extension of SEARCH that gives access to advanced functionalities of Twitter, such as to seek users, followers, to see the status, to publish information on its account, etc.

STREAMING: allows communicating with Twitter in streaming mode. The advanced feature of the API is that it allows access to large volume of Twitter data and to be less constraining by the limited access in interrogation mode. However, this API requires the installation of a more complex mechanism to access the data.

WEBSITES: allow integrating basic functions of the Twitter in websites.

Twitter is an information network and a means of communication that produces more than 200 million tweets per day. The Twitter platform offers access to this corpus of data via the API. Each API represents a facet of Twitter, allowing the developers to access the Twitter database.

*Twitter for Web*

Twitter for websites (TfW) allows the integration and facilitates the installation of Twitter on the web. TfW is ideal for site developers and webmasters to integrate easily the basic functionalities for Twitter in their website. This includes offers like the tweet button, which enables a user to tweet something that he/she found engaging on the website to all the friends. The friends see the tweet go through the website, and a certain percentage of the tweeters share the same contents with their friends, thus forming an outer loop of distribution.

In the same way, the 'Follow' button makes it possible for a user to follow an account in the background on Twitter. This particular user will then get updates of the account directed in the background of his current profile, creating a new sequence of engagement with the user base.

We will cover subjects in various fields, such as the election, football, racism and freedom.

We used the algorithms Naive Bayes and Turney, which are based on the opinion mining in order to detect the opinions of the net surfers.

TREATMENT PROCESS

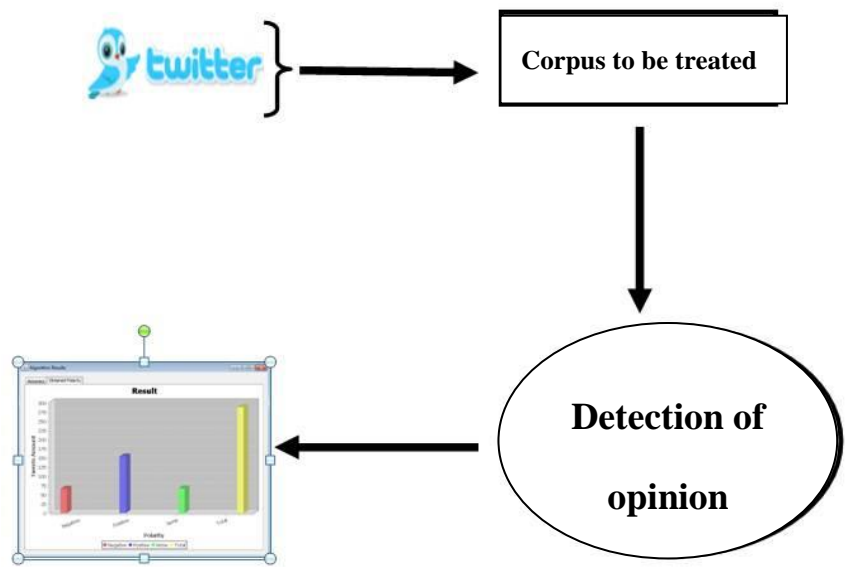

**Figure 2**: Process and treatment of the opinion detection

(1) *Corpus to be treated* -- This gathers the tweets on Twitter based on the specified research request and the filters inside a specified set of dates.

(2) *Detection of opinion* -- This takes the tweets that are joined together and analyses them based on their feeling for each given request.

SOME EXAMPLES OF COLLECTED CORPUS

*Corpus of the Egypt Elections*

> The Muslim brotherhood is winning the **election** with 40% in **Egypt** and 2nd place is the ultra conservative Islamic party **:(** #pray4copts
> For the 1st time in history, **Egyptians** will not know the result of an **election** b4 they head to the polls to vote **:)** Go **Egypt**! #Egy**Elections**
> happy hard presidential **election** day in #**egypt** may god give us the right leader to prosper and save our nation **:)**
> **Egypt** holding first ever free presidential **election.** Democracy is a beautiful thing **:)** #arabspring #powertothepeople

*Corpus on the Social Networks*

> Someone text or **social network** me, boring people are going to sleep **:(**
> Can't SMS, u seldom online **social network** ... How ? I want to spend time yet u like avoiding or is it I am hesitating?**:(**
> the twitter without u is just one more **social network** without fun **:(**
> Mid-year exams starts next week ..sigh #nodistractionsplease ..I am banning myself from all **social network** ............**:(**
> Twitter is supposed to be a **social network** for gods sake, not somewhere where you can bully people and have an I know best attitude. **:(**

*Corpus on Immigration*

> Ms. Z- "What is divergent evolution?" Student- "is it like **immigration**?" LMAO!!!! Welcome to Riverside **:P**
> I'm very pro-**immigration**, every ethnic group that enters the country takes more heat off black guys **:D**
> Obama swore on fixing the **immigration** issue and now what I hate a niggah he doesn't have my vote **:-)**
> Sending POWERFUL vibes to my King today who has a very important **immigration** hearing at 2:30...your thoughts are appreciated. <3 **:)**
> Thru **immigration** baggage claim customs rechecked bags security...hello d terminal **:)**

*Corpus on UEFA (Union of European Football Associations)*

> wellwellwell... so it begins in 2 weeks. **UEFA** EURO 2012! Can't wait! **:D**
> So the 2012 **UEFA** Euro League starts next month... Germany has the most victories with 3 titles. Let's kick ass in 2012 and make it 4! **:D**
> Just watched the highlights of **UEFA** Final ... Seriously it wasn't the day for Bayern Munich. **:P**
> congratulations to chelsea fc for winning the 2012 **uefa** champions league **:D**
> sayanglangtalagaang real madrid at barcelona**:D**
> I wanna congratulate the Champions of Europe **:)** this was long due. If you are a great fan of #Chelsea. follow me and I follow back #**UEFA**

We have collected 20 corpora by using the API Twitter4J because with this API, we can collect the corpora on current events' subjects.

- Examples of emoticons used in our study are as follows:
- Emoticons happy: ":-)", ":)", "=)", ":D", etc.
- Emoticons sad: ":-(", ":(", "= (", ";(", etc.

RESULTS

We have tested some corpora from which we have extracted the tweets and then we have applied the Bayesian probabilistic algorithm and also the Turney's algorithm. Results are listed in the following Tables 2 and 3.

**Table 2**: Detection of opinions by Bayesian algorithm

| Corpus name | #Tweets | #Positive tweets | #Negative tweets | #Neutral tweets | Process time (s) |
|---|---|---|---|---|---|
| Book | 290 | 197 | 93 | 0 | 20 |
| Cinema | 96 | 71 | 25 | 0 | 22 |
| Cuisine | 94 | 64 | 29 | 0 | 34 |
| Education | 137 | 75 | 61 | 1 | 19 |
| Election | 220 | 117 | 103 | 0 | 31 |
| Facebook | 56 | 38 | 18 | 0 | 13 |
| Football | 291 | 162 | 129 | 0 | 23 |
| Freedom | 191 | 96 | 95 | 0 | 36 |
| Google | 10113 | 6823 | 3274 | 14 | 64 |
| Internet | 346 | 226 | 117 | 3 | 12 |
| Movie | 740 | 543 | 197 | 0 | 50 |
| Music | 34 | 22 | 11 | 10 | 43 |
| Politic | 132 | 77 | 54 | 10 | 50 |
| Racism | 76 | 45 | 31 | 0 | 37 |
| Religion | 293 | 159 | 134 | 0 | 35 |
| Social network | 173 | 140 | 33 | 0 | 30 |
| Sport | 105 | 55 | 50 | 0 | 28 |
| Travel | 142 | 99 | 43 | 0 | 25 |
| Video game | 77 | 54 | 23 | 0 | 25 |
| Violence | 222 | 114 | 108 | 0 | 36 |

**Table 3:** Detection of opinions by Turney's technique

| Corpus name | #Tweets | #Positive tweets | #Negative tweets | #Neutral tweets | Process time (s) |
|---|---|---|---|---|---|
| Book | 290 | 155 | 67 | 68 | 140 |
| Cinema | 96 | 57 | 22 | 17 | 142 |
| Cuisine | 94 | 66 | 12 | 16 | 154 |
| Education | 137 | 118 | 19 | 0 | 139 |
| Election | 220 | 95 | 64 | 61 | 151 |
| Facebook | 56 | 32 | 20 | 4 | 133 |
| Football | 291 | 132 | 110 | 49 | 143 |
| Freedom | 191 | 141 | 33 | 17 | 156 |
| Google | 10113 | 4051 | 1558 | 4454 | 184 |
| Internet | 346 | 193 | 90 | 63 | 132 |
| Movie | 740 | 384 | 216 | 140 | 170 |
| Music | 34 | 15 | 9 | 10 | 163 |
| Politic | 132 | 67 | 41 | 24 | 170 |
| Racism | 76 | 20 | 52 | 4 | 157 |
| Religion | 293 | 122 | 93 | 78 | 155 |
| Social network | 173 | 57 | 72 | 44 | 150 |
| Sport | 105 | 64 | 20 | 21 | 148 |
| Travel | 142 | 93 | 26 | 23 | 145 |
| Video game | 77 | 33 | 36 | 8 | 145 |
| Violence | 222 | 82 | 106 | 32 | 156 |

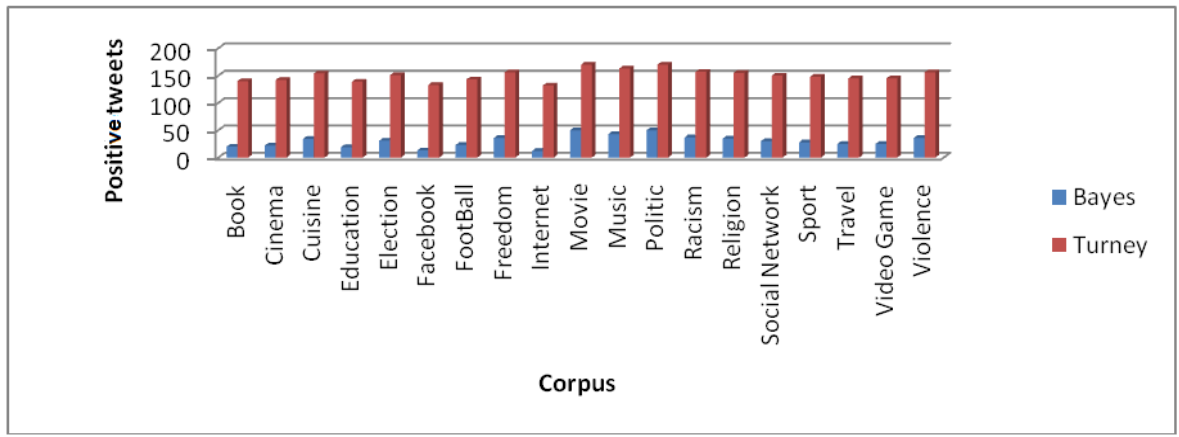

**Figure 3**: Graphical representation of positive tweets for Bayesian techniques and Turney's algorithm.

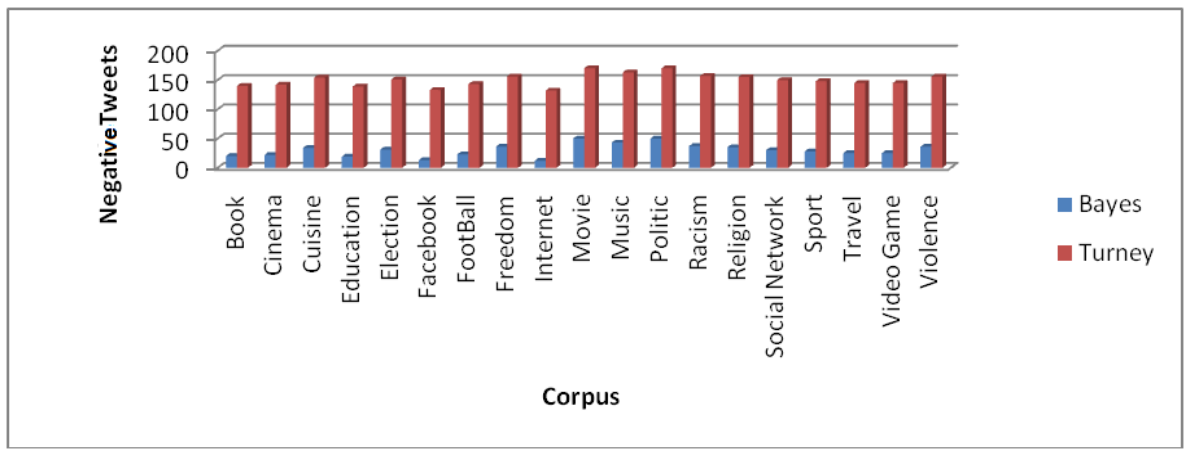

**Figure 4**: Graphical representation of negative tweets for Bayesian techniques and Turney's algorithm.

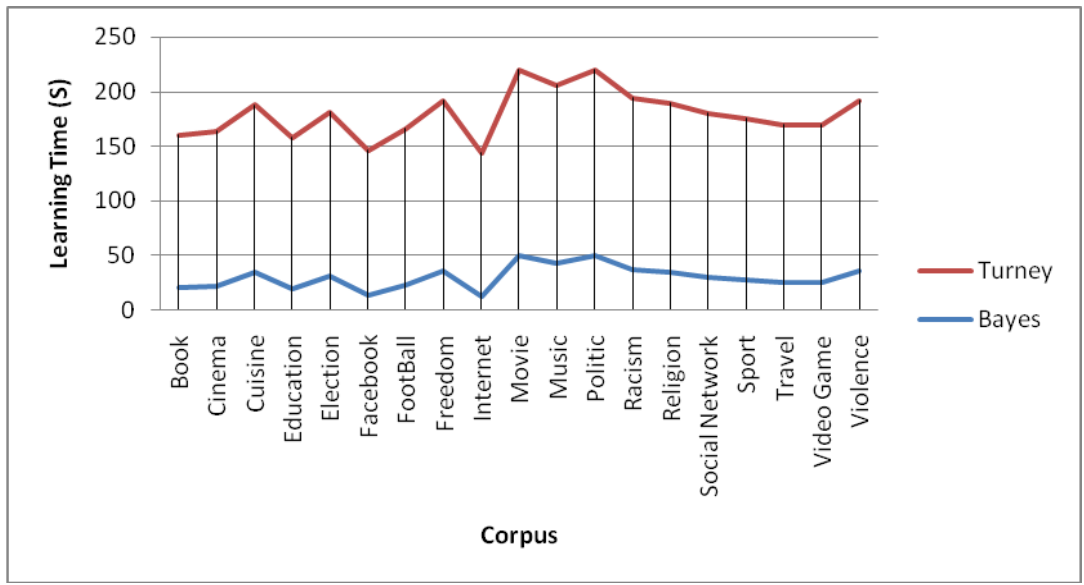

**Figure 5**: Graphical representation of learning time for Bayesian techniques and Turney's algorithm.

We notice that the computing time of the method of Turney is significantly longer than that of the method of Bayes because the first one carries out a research on Internet but the second calls upon the local training set.

We also notice that it has a great difference in the number of the neutral tweets, which is caused by the sentences that do not contain adjectives and adverbs, but which contain emoticons (either positive or negative).

**CONCLUSION AND PERSPECTIVES**

In this work, we have presented two approaches for the detection of opinions in tweets of Twitter -- a supervised learning approach and a SO approach; the first one is based on the emoticons, whereas the second is based on adjectives.

The experiments carried out have led to positive results and many perspectives are yet to be explored in this rapidly expanding field.